\def\year{2020}\relax
\documentclass[letterpaper]{article} 
\usepackage{aaai20}  
\usepackage{times}  
\usepackage{helvet} 
\usepackage{courier}  
\usepackage[hyphens]{url}  
\usepackage{graphicx} 
\usepackage{graphicx}  
\usepackage{amsmath}
\usepackage{amssymb}
\newcommand{\argmax}{\mathop{\mathrm{argmax}}\limits} 
\usepackage{enumitem}

\usepackage{subcaption}
\usepackage[acronym]{glossaries}
\usepackage{sidecap}
\usepackage{tabularx, booktabs, ragged2e}
\usepackage{adjustbox}
\usepackage{multirow}
\usepackage{makecell}

\title{Joint Super-Resolution and Alignment of Tiny Faces
}

\author{Yu Yin,\textsuperscript{\rm 1}
Joseph P. Robinson,\textsuperscript{\rm 1}
Yulun Zhang,\textsuperscript{\rm 1}
Yun Fu\textsuperscript{\rm 1, 2}\\
\textsuperscript{\rm 1}Department of Electrical and Computer Engineering, Northeastern University, Boston, MA\\
\textsuperscript{\rm 2}Khoury College of Computer Science, Northeastern University, Boston, MA\\
\{yin.yu1, robinson.jo\}@husky.neu.edu, yulun100@gmail.com, yunfu@ece.neu.edu
}
 
\begin{document}
\newcommand{\ie}{\textit{i}.\textit{e}., }
\newcommand{\eg}{\textit{e}.\textit{g}., }
\newcommand*{\etc}{etc.\@\xspace}%

\maketitle

\begin{abstract}
Super-resolution (SR) and landmark localization of tiny faces are highly correlated tasks. On the one hand, landmark localization could obtain higher accuracy with faces of high-resolution (HR). On the other hand, face SR would benefit from prior knowledge of facial attributes such as landmarks. 
Thus, we propose a joint alignment and SR network to simultaneously detect facial landmarks and super-resolve tiny faces. More specifically, a shared deep encoder is applied to extract features for both tasks by leveraging complementary information. To exploit representative power of the hierarchical encoder, intermediate layers of a shared feature extraction module are fused to form efficient feature representations.
The fused features are then fed to task-specific modules to detect landmarks and super-resolve face images in parallel.
Extensive experiments demonstrate that the proposed model significantly outperforms the state-of-the-art in both landmark localization and SR of faces. We show a large improvement for landmark localization of tiny faces (\ie $16\times16$). Furthermore, the proposed framework yields comparable results for landmark localization on low-resolution (LR) faces (\ie $64\times64$) to existing methods on HR (\ie $256\times256$). As for SR, the proposed method recovers sharper edges and more details from LR face images than other state-of-the-art methods, which we demonstrate qualitatively and quantitatively.
\end{abstract}

\begin{figure}[t]
 \centering
    \includegraphics[width=0.95\linewidth]{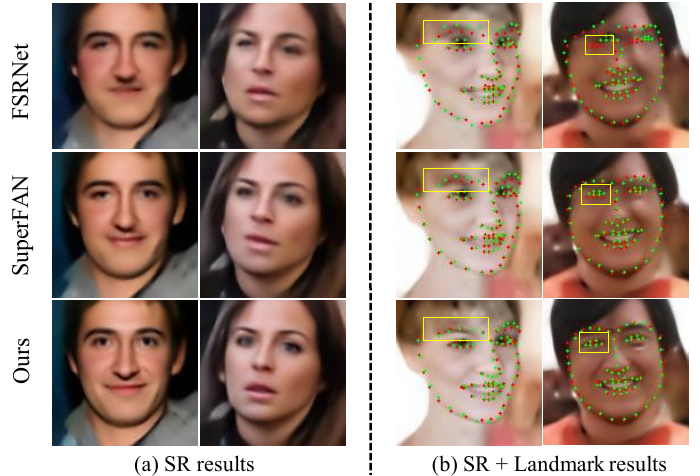}
    \caption{\textbf{Comparison with SuperFAN (Bulat et al. 2018) and FSRNet \cite{chen2018fsrnet}.} The proposed recovers sharper edges and finer details HR space (a). Also, (b) shows estimated landmarks superimposed on SR face, where Red marks true landmarks and green marks predicted landmarks.}
\label{fig:resultsOnBothTasks}
\end{figure}

\section{Introduction}
Automatic face understanding is critical for problems in human perception (\eg super-resolution (SR)~\cite{yu2016ultra}, visual understanding~\cite{guccluturk2017reconstructing}, and style transfer~\cite{liu2017unsupervised}) and applied machine vision (\eg landmark localization~\cite{robinson2019laplace}, identity recognition~\cite{wu2016deep}, and face detection~\cite{zhang2016joint}). Modern-day models for face-based tasks tend to breakdown when applied to images of low-resolution (LR). In practice, face-based systems are frequently confronted with such scenarios (\eg LR cameras used for surveillance~\cite{yu2017hallucinating}). Recent studies revealed that a decrease in resolution (\ie $<30\times30$) yields an increase in error for models used for facial landmark localization (Bulat et al. 2018). To address this problem, face SR, also known as face hallucination, aims to generate high-resolution (HR) faces from LR imagery~\cite{Liu2007}. The recovered faces then provide more detailed information (\eg sharper edges, clearer shapes, and finer skin details), and are often used for improved analysis and perception. However, most existing methods (\eg Superfan~\cite{bulat2018superfan}) rely heavily on the quality of recovered images. Since SR methods usually suffer from blurriness, using SR images for face-related tasks can hinder the final prediction or conclusion.

On the other hand, facial prior knowledge can be used to recover SR faces of higher quality~\cite{Baker2000,Liu2007}. In problems of single image super-resolution (SISR), face SR utilizes prior knowledge to improve the accuracy of the inferred images and, thus, to yield results of higher quality. For example, one can leverage low-level information (\ie smoothness in color), facial heatmaps, and face parsing maps to provide additional mid-level information (\ie face structure) to recover sharper edges and shapes~\cite{chen2018fsrnet}. Also, high-level information can be extracted with identity labels and other face attributes (\eg gender, age, and pose), and then leveraged to reduce the ambiguity of the hallucinated faces~\cite{yu2018super,lee2018attribute}. Hence, additional face information is beneficial for SR, and especially for tiny faces (\eg $16\times16$).

\begin{figure}[t]
 \centering
    \includegraphics[width=\linewidth]{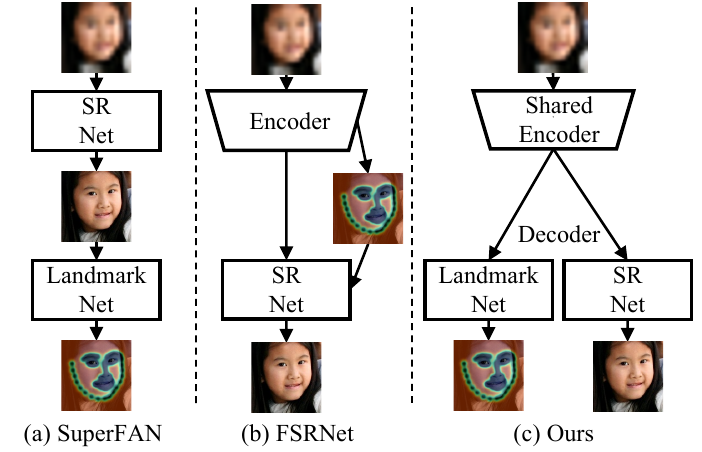}
\caption{\textbf{Graphical view.} (a) SuperFAN~\cite{bulat2018superfan} detects landmarks on super-resolved faces. (b) FSRNet~\cite{chen2018fsrnet} uses prior information for SR. (c) Our multi-task framework jointly learns landmark localization and SR, with tasks aiding one another.}
\label{fig:frameworkComp}
\end{figure}

Previous work in face SR either super-resolved LR images using prior information (\eg FSRNet~\cite{chen2018fsrnet}) or directly localized the landmarks on the super-resolved images (\eg SuperFAN~\cite{bulat2018superfan}). Figure \ref{fig:frameworkComp} compares these frameworks with the proposed method. 
Specifically, SuperFAN only uses SR to help localize the landmarks of tiny faces, but not vice-versa. Besides, our model does not process the recovered SR output that suffers from blurriness, as we dedicate an encoding module to maximize the amount of information captured from LR faces. As for FSRNet, landmarks are only used as facial prior knowledge to super-resolve faces, which suffers from the same problem of detecting landmarks on a coarse, recovered SR image.
Furthermore, SuperFAN and FSRNet address the two tasks separately, leading to redundant feature maps.
Since both face SR and landmark localization tasks could benefit from one another, we aim to extract the maximum amount of information from LR faces by addressing the two tasks simultaneously.
Thus, we propose a multi-task framework that allows these tasks to benefit from one another, which improves the performance in both tasks (see Figure\ref{fig:resultsOnBothTasks}). 

The main contribution of this paper are as follows: 
\begin{enumerate}
    \item In this paper, we propose a network that does SR and landmark detection on tiny faces jointly-- a network we dubbed JASRNet\footnote{The code is available at: https://github.com/YuYin1/JASRNet.}. To the best of our knowledge, we are the first to train a multi-task model  that jointly learns landmark localization and SR. Specifically, and unlike existing two-step approaches, we leverage the complementary information of the two tasks. This allows for more accurate landmark predictions to be made in LR space and improved reconstruction from LR-to-HR.
    \item Novel deep feature extraction and fusion modules are used to maximize the amount of information captured from the LR faces, which is done at intermediate layers of the encoder to exploit the deep hierarchical machinery. 
    \item We show large improvements for both SR and landmark localization for tiny faces (\ie $16\times16$). Besides, our JASRNet yields results for landmark localization on LR faces (\ie $64\times64$) that are comparable to existing methods evaluated on the corresponding HR faces (\ie $256\times256$). Furthermore, the proposed method recovers HR faces with sharper edges and shapes compared with state-of-the-art methods for SR. 

\end{enumerate}

\section{Related Work}
\subsection{Face super-resolution}
Typical SISR methods do not benefit from facial prior information and can be utilized to super-resolve images of arbitrary type. By introducing face-specific information, Yu~\cite{yu2016ultra,yu2017hallucinating} proposed a GAN-based model to recover HR images from tiny faces of size $16\times16$. Chen~\cite{chen2018fsrnet} used a separate branch to estimate facial landmark heatmaps and parsing maps, which were then used as face-specific information to super-resolve tiny face images. FaceAttr~\cite{yu2018super} validated that knowledge of facial attributes can also significantly reduce the ambiguity in face SR. It is worth noting that our method not only utilizes facial prior information to super-resolve tiny face with better quality, but also achieves state-of-the-art performance on landmark alignment by benefiting from SR.

\subsection{Face alignment}
Modern-day approaches for face alignment have been successful on HR faces~\cite{dong2018supervision,lv2017deep,mo2019face,ranjan2019hyperface} . 
However, most suffer from performance degradation with decreasing image resolution, especially with faces smaller than $30\times30$~\cite{bulat2017far}.
The first to address landmark detection on LR faces was SuperFAN~\cite{bulat2018superfan}, which super-resolved tiny faces, from which the output images were fed to a landmark localization model. Although the error of the landmark localization provides gradients to back-propagate through the SR module, it is, in essence, a 2-step process. We argue that the facial prior information is not fully utilized for SR. To address this problem, we present a novel synergistic multi-task framework that learns facial landmark localization and SR jointly.

\begin{figure*}[t]
 \centering
    \includegraphics[trim=0in 0in 0in 0in,clip,width=.95\linewidth]{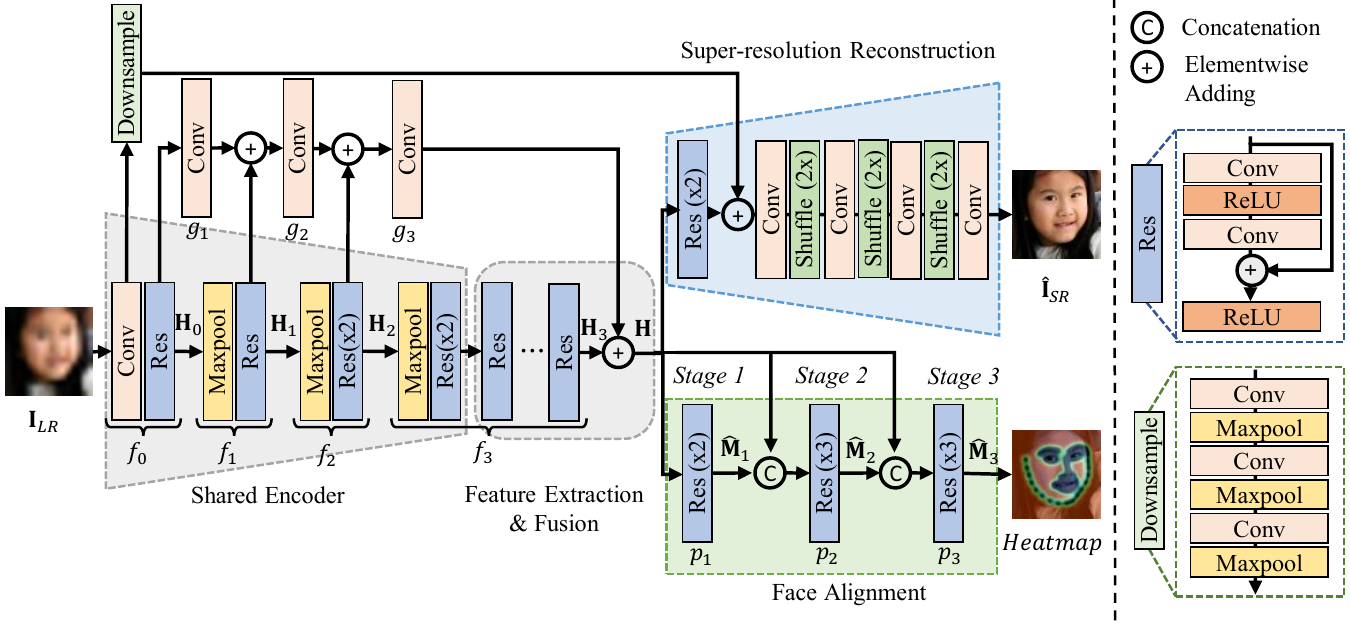}
\caption{\textbf{Architecture of the proposed JASRNet.} The shared encoder module is used for extracting shallow and shared features for both tasks. The deep feature extraction and fusion module is used for obtaining better feature representations. The other two modules are task-specific modules for super-resolution and face alignment, respectively.
}
\label{fig:netStructure}
\end{figure*}

\subsection{Multi-task learning}
Multi-task learning is commonly used to jointly address correlated tasks. HyperFace~\cite{ranjan2019hyperface} proposed a multi-task learning framework for face detection, face alignment, gender recognition, and pose estimation. The joint learning tasks were based on regression or classification (\ie a special case of regression). Hence, similar architectures were adopted for all tasks. In our case, however, face SR and alignment are based on generation and regression, respectively. Thus, one of the main differences in architectures of the proposed from HyperFace is that we include specific modules for each task, while HyperFace used only fully connected layers after feature fusion.

\section{Method}
Super-resolution (SR) and landmark localization of tiny faces are highly correlated tasks. Both of them can benefit from each other. While previous work either uses SR to help align tiny faces or vice-versa, but not both. We argue that the amount of information extracted from LR image is not maximized when only one task is used to help the other.
Hence, we propose a deep joint alignment and super-resolution network (JASRNet) to model super-resolution and localize landmarks for tiny faces simultaneously, with information from both tasks boosting the performance of the other. As shown in Figure \ref{fig:netStructure}, the proposed JASRNet consists of four parts: (1) a shared shallow encoder module is used for extracting shallow and shared features for both tasks; (2) a deep feature extraction and fusion, which is used for obtaining better feature representations; (3-4) task-specific modules for super-resolution and face alignment, respectively.

Let $\{\mathbf{I}^{(i)}_{LR}, \mathbf{I}^{(i)}_{HR}, \mathbf{M}^{(i)}\}^N_{i=1}$ be $N$ training samples. The original LR faces $\mathbf{I}^{(i)}_{LR}$ are passed in to the shared encoder, which then feeds into the feature extraction module to extract features for both tasks. 
To exploit the representative power of different grains,
the intermediate features of the shared encoder branch out to fuse with the output of deep feature extraction module. This feature fusion forms a more efficient feature representation, as later demonstrated as part of the ablation study. Carrying on, the fused features are fed to both task-specific modules. Thus, the super-resolved images $\mathbf{\hat{I}}^{(i)}_{SR}$ and the probability maps of the landmark estimations $\mathbf{\hat{M}}^{(i)}$ are produced simultaneously. 

Usually, there are sharper edges or sudden changes around the contour of facial component. For face alignment, the SR module recovers the image with better resolution, which, hence, helps the model detect more accurate landmarks. In parallel, the alignment module locates the edges and structure of the face, forcing more attention to the high-frequency content (\ie edges).
Since both tasks, face SR and landmark localization, are suited to benefit from one another, the aim of this work is to exploit the amount of maximum information that can be extracted from the LR faces. This is done by combining the loss function of each task.
For the SR task, the $L_1$ loss is minimized, as it can provide better convergence than $L_2$ \cite{lim2017enhanced,zhang2018residual}. 
For the alignment task, a $L_2$ heatmap loss is used, like in~\cite{dong2018supervision}. Together, the loss function of JASRNet can be expressed as
\begin{equation}
\begin{split}
l & = l_{sr} + \alpha l_{heatmap} \\
 & = \frac{1}{N}\sum^N_{i=1}\{\| \mathbf{\hat{I}}^{(i)}_{SR} - \mathbf{I}^{(i)}_{HR} \|_1 + \alpha\| \mathbf{\hat{M}}^{(i)} - \mathbf{M}^{(i)} \|_2\},
\end{split}
\end{equation}
where $l$ denotes the total loss, $l_{sr}$ and $l_{heatmap}$ denote the $L_1$ loss for super-resolution and the $L_2$ heatmap loss for alignment, respectively.
The weight of $l_{heatmap}$ is $\alpha$, and the estimated heatmap of the $i^{th}$ image is $\mathbf{\hat{M}}^{(i)}$. As mentioned above, $\mathbf{\hat{I}}^{(i)}_{SR}$ is the super-resolved image recovered from $\mathbf{I}^{(i)}_{LR}$.


\subsection{Shared feature extraction and fusion}
\noindent\textbf{Shallow encoder}.
Previous work in face SR and alignment usually addressed these two tasks separately, leading to redundant feature maps. To efficiently extract features from LR images, a shared encoder is designed to extract shallow features that capture complementary information of the two tasks. It consists of a convolutional layer, a residual block~\cite{he2016deep}, and then three transformations made-up of the maxpooling operation and residual blocks (Figure~\ref{fig:netStructure}). Intermediate layers of the encoder are later fused for richer features in geometry and semantics.

All the convolution layers of JASRNet use kernels of size $3\times3$, and each is followed by a ReLU layer. The number of channels are all set as 128, except for the last convolutional layers in both reconstruction and alignment module, which are set as 3 and the number of landmarks (namely 68 for 300W), respectively.
There are three maxpooling layers in the network, each downsample the feature maps $\times2$, which, in total, reduce the size of the feature maps by a factor of 8. The structure of the residual blocks is the same as in the original residual nets (ResNets)~\cite{he2016deep}, except we omitted the batch normalization (BN) layers, as it reduces the variation of feature ranges: ResNets used for SISR (EDSR) performed best with all BN layers removed~\cite{lim2017enhanced}.
Also, we found that BN layers slow down the speed to convergence of the network, while reducing its overall performance, which was especially true in the SR task. Since we aim to reserve the most information possible when passing through the shared encoder module (\ie during feature extraction), we follow EDSR~\cite{lim2017enhanced} and remove all BN from the residual blocks. 

\noindent\textbf{Deep feature extraction and fusion}.
\label{sec:deepFeature}
Deeper networks have shown to have a better performance in many computer vision tasks including SR \cite{bulat2018superfan,chen2018fsrnet,he2016deep,lim2017enhanced,tai2017memnet}. Increased depth was also a tactic used in this work.
Shallow features extracted from the shared encoder are passed to the deep feature extraction module consisting of $T$ residual blocks, with $T=32$ in the reported experiments.
A deeper network not only recovers sharper edges and shapes for super-resolved face images, but it also achieves a higher accuracy for landmark localization.


Inspired by Hyperface~\cite{ranjan2019hyperface}, we fused intermediate layers to exploit the representative power of features at different levels of the hierarchical model. Considering the similarity of features from adjacent layers, not all features of the shared encoder are fused to compose the new feature representation. Since each of the maxpooling layers downsample the feature map by a factor of 2, the output of the layers that precede each maxpooling layer branches out using skip connections, and are later fused to form richer features with geometry information. To match sizes of the feature maps, a $3\times3$ convolutional layer with stride 2 is applied to downsample fusing features by a factor of 2 for each maxpooling layer that is applied in parallel to the skip connection.

The outputs before the maxpooling are denoted as $\mathbf{H}_i  (i \in \{0, 1, 2\})$; the output of the last residual block in feature extraction module is $\mathbf{H}_3$ (see Figure \ref{fig:netStructure}). Provided LR images $\mathbf{I}_{LR}$ as input, we have
\begin{gather*} 
\mathbf{H}_0  = f_0(\mathbf{I}_{LR}), \\ 
\mathbf{H}_i = f_i(\mathbf{H}_{(i-1)}), i \in \{1, 2, 3\},
\end{gather*}
where $f_i(\cdot)$ $(i \in \{0, 1, 2, 3\})$ transform the signal during feature extraction. Hence, $f_0$ is the mapping of the first convolution layers and residual blocks, $f_1$ and $f_2$ are the mappings of the first and second steps combining maxpooling and residual blocks, respectively, and $f_3$ is the mapping for the remaining residual blocks making up the feature extraction module.

Mathematically speaking, the fused features $\mathbf{H}$ that is output can be founded as
\begin{equation}
\mathbf{H} = g_3( g_2( g_1(\mathbf{H}_0) + \mathbf{H}_1) + \mathbf{H}_2) + \mathbf{H}_3,
\end{equation}
where the convolution operation $g_i(\cdot)$ $(i \in \{1, 2, 3\})$ fuses intermediate features.

\subsection{Task-specific modules}
\label{sec:taskSpecific}
\noindent\textbf{Super-resolution reconstruction}.
The \textit{super-resolution reconstruction} module reconstructs the HR image from shared features of size $16\times16$. First, shared feature maps are fed to two residual blocks to extract task-specific features. 
Next, 3 conv-layers, each of which are followed by pixel shuffle layers~\cite{shi2016real}, upscale the feature maps $2\times$ in size (\ie $16\times16$ to $128\times128$). Finally, a convolutional layer made-up of $3\times3$ filters to map from HR RGB image space.

Inspired by EDSR~\cite{lim2017enhanced} and RDN~\cite{zhang2018residual}, the first and last residual blocks of the shared encoder and SR reconstruction module are linked by a large skip connection. This recovers HR images with finer details (\ie sharper edges and shapes). The skip connection directly provides low frequency information to the super-resolved images. Hence, it forces the network to focus on learning the high frequency information, opposed to low frequency information already provided.
Since the output size of the first convolution layer is $128\times128$, and the feature map size of the last residual block in reconstruction module is $16\times16$, we downsample the $128\times128$ feature map $8\times$ with 3 convolution and 3 maxpooling layers (see Figure \ref{fig:netStructure}). 

Unlike SuperFAN, where the long skip connection is reported to have minimal impact on overall performance, our model largely benefits from the skip connection. This is because the features extracted includes high frequency information and, thus, is more efficient for recovering sharp and accurate edges. Furthermore, since super-resolution and face alignment share the deep features, a byproduct of this long skip connection also is boosted performance for the landmark localization task as well.

\begin{figure*}[t]
 \centering
\includegraphics[width=0.95\linewidth]{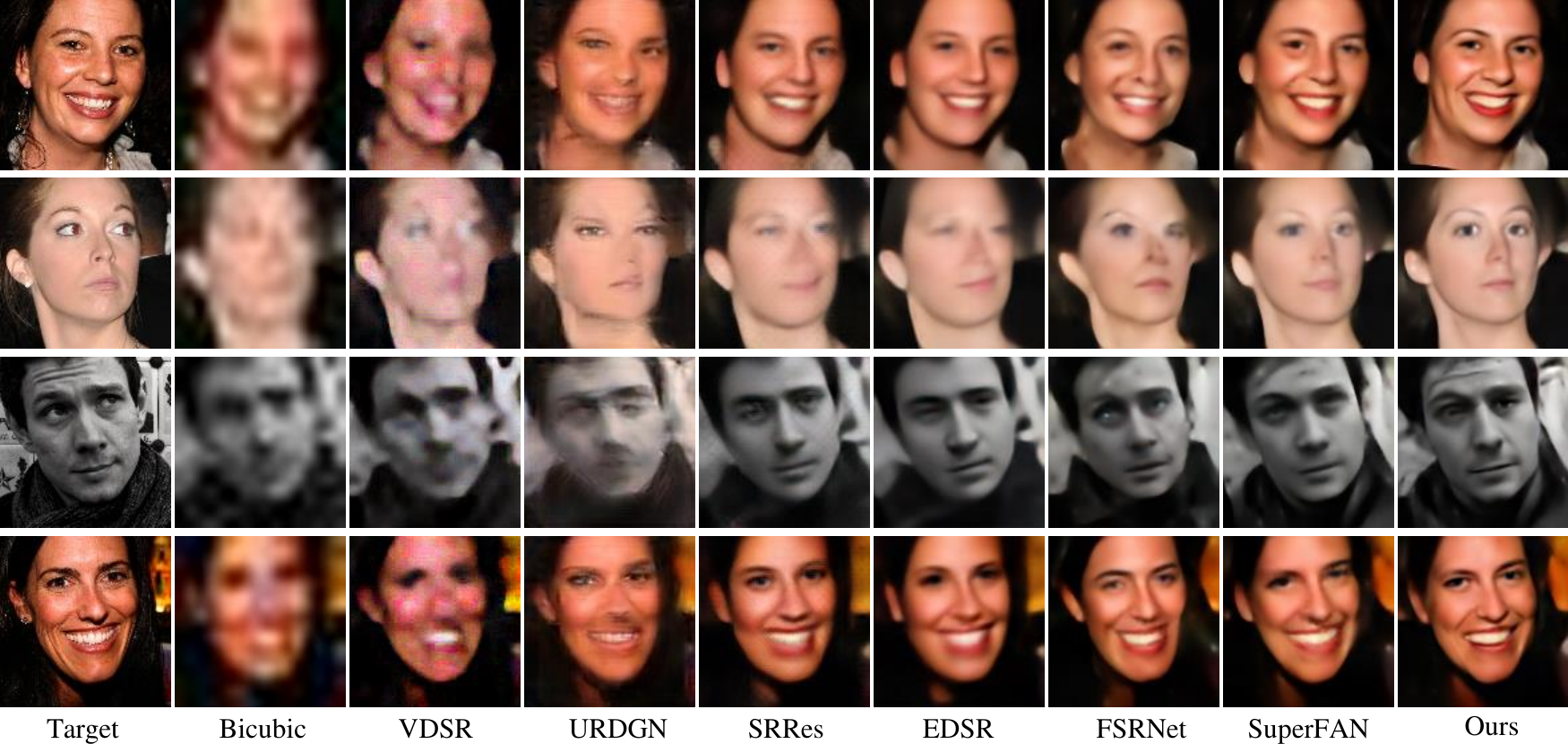}
\caption{\textbf{Visual results.} Comparison of different super-resolution methods.}
\label{fig:SRsmpImg}
\end{figure*}

\begin{table*}[t]
    \caption{\textbf{Quantitative comparisons.} PSNR/SSIM on 300W and HELEN.}
    \centering
    {\footnotesize
    \begin{adjustbox}{max width=\linewidth}
        \begin{tabular}{ c  ccccccccc}
            \toprule
             & Bicubic & VDSR & URDGN & SRRes & EDSR & TDAE & FSRNet & SuperFAN & Ours\\\midrule 
            300W & 21.36/0.594 & 21.80/0.558 & 21.97/0.617 & 23.30/0.669 & 23.47/0.658 & 21.12/0.547 & 23.05/0.678 & 23.13/0.691 & \textbf{23.69}/\textbf{0.711}\\
            HELEN & 21.36/0.593 & 21.66/0.552 & 21.77/0.605 & 23.05/0.674 & 23.40/0.709&  21.70/0.542 & - - / - - & 23.17/0.695 & \textbf{23.55}/\textbf{0.717}\\ \bottomrule
        \end{tabular}
    \label{tab:resultsSR}
    \end{adjustbox}
    }
\end{table*}

\noindent\textbf{Face alignment}.
Like the \textit{SR reconstruction} module, the shared features $\mathbf{H}$ are fed through consecutive residual blocks to extract features specific to face alignment. 
Inspired by the successes of convolutional pose machines (CPM) ~\cite{wei2016convolutional} on face alignment, we also utilize the sequential framework made-up of residual blocks for estimating locations of landmarks. In the first stage, two residual blocks predict coarse heatmaps $\mathbf{\hat{M}_1}$. Then, in the second stage, the heatmaps $\mathbf{\hat{M}_1}$ predicted in the first stage are first concatenated with the feature maps $\mathbf{H}$, which are then fed to the second prediction module composed of three sequential residual blocks that predict heatmaps $\mathbf{\hat{M}_2}$. The third stage then concatenates feature maps $\mathbf{H}$ and $\mathbf{\hat{M}_2}$ to produce final estimation $\mathbf{\hat{M}_3}$ expressed as
\begin{equation}
\mathbf{\hat{M}_3} = p_3([\mathbf{H}, p_2([\mathbf{H},  p_1(\mathbf{H})])]),
\end{equation}
where $p_j$ maps the prediction modules, with $(j \in \{1, 2, 3\})$.
Note that the size of the feature maps is constant throughout the face alignment module (\ie $16\times16$). During training, heatmap regression $L_2$ loss was used to localize landmarks, opposed to directly predicting pixel coordinates $(x, y)$. Thus, argmax is used to determine $(x, y)$ from the predicted heatmaps in final stage (\ie $\hat{M}_3$). Specifically, the maximum value of each of the $K$ heatmaps is found as the predicted landmarks (\ie $\argmax_{(x, y)}\hat{M}_3$).

\section{Experiments}\label{sec:experiments}
We now review the experimental settings and results. Specifically, the datasets, implementation details, and metrics are first described. 
Then, we show results comparing with the state-of-the-art methods for the face SR and alignment task separately.
Besides, we highlight the benefits of the proposed feature fusion and joint training. Finally, we conduct an ablation study as a deep-dive revealing the contributions of the components introduced in this work.

\subsection{Experimental settings}\label{sec:experimentsSetting}
\noindent\textbf{Datasets}. We evaluated the proposed approach on several datasets, which are listed as follows:
\begin{itemize}[leftmargin=*]
\setlength{\itemindent}{0em}
    \item \textbf{300W}~\cite{sagonas2013300,sagonas2016300} consists of 3,837 face images with 68 landmarks. We used the same training set as ~\cite{lv2017deep,zhu2015face}. Subsets of 300W were evaluated: \textit{common} and \textit{challenge}, and \textit{full}. 
    \item \textbf{AFLW}~\cite{koestinger2011annotated} consists of 24,386 faces, each with 21 landmarks. The dataset was split into 20,000 faces for training and the rest (\ie 4,386) for testing ~\cite{dong2018supervision}. Also, the left and right ears were ignored, leaving up to 19 landmarks per face sample.
    \item \textbf{HELEN}~\cite{le2012interactive} contains 2,330 images. The annotation of all 194 landmarks were used as facial prior information. We followed~\cite{chen2018fsrnet} to use the last 50 images for testing and the rest for training.
    \item \textbf{LFW}~\cite{LFWTech,LFWTechUpdate} contains 13,233 face images collected from 5,750 people. Each image is labeled with the name of the person pictured. Hence, it will also be used to evaluate the recognition capabilities of super-resolved images. Note that this dataset was only used for testing.
\end{itemize}

\noindent\textbf{Implementation details}. We first cropped facial images about the head region, which were then resized to 128$\times$128. These were designated as the HR images. Then, LR images were generated by applying bicubic downsampling (8$\times$) to the HR images, yielding a resolution of 16$\times$16. Then, the input LR images were reversed to match the size of the HR faces: each were up-scaled 8$\times$ using bicubic interpolation resulting in images of size 128$\times$128. The training images were augmented using random scaling, rotation, and horizontally flipping. Specifically, these augmentation transformations were used to make fifteen copies.
Optimization was done with ADAM with a learning rate of $5.0\times10^{-5}$ that dropped 0.5 at $20^{th}$ and $30^{th}$ epochs. The model was trained with a batch size of 8 and for a total epoch of 40 epochs.
Implementation was done using PyTorch. Training took about 7 hours on Helen with a Nvidia TITAN-XP GPU.

\noindent\textbf{Evaluation metrics}. The metric used to evaluate landmark localization was NMSE (\ie the normalized euclidean distances between ground-truth and predicted landmarks). Following ~\cite{bulat2017far,dong2018supervision,sagonas2013300}, the normalization factor is set as inter-ocular distance for 300W and the area of the ground-truth bounding box for AFLW dataset.

For SR, we evaluated using the peak signal to noise ratio (PSNR) and structural similarity index (SSIM)~\cite{wang2004image}: PSNR is computed as the mean squared error (MSE) between the SR and HR images, while SSIM accounts for the noise and edges  (\ie the high-frequency content) of an image.
In our experiments, we converted the RGB images to the YCbCr color space and only calculated the PSNR for the Y-channel. To focus on the face region, while ignoring the background, only the face region within the bounding box was measured when evaluating the SR images.

\subsection{Comparison with state-of-the-art methods}
Comparisons were made with state-of-the-art methods in both SR and face alignment. It is important to note that most existing methods only do a single task, while the proposed model does both. Furthermore, our model performs the best in both tasks. The methods that do both tasks, SuperFAN~\cite{bulat2018superfan} and FSRNet~\cite{chen2018fsrnet}, were used to compare both tasks simultaneously.

\begin{table}[t!]
    \caption{\textbf{NMSE on 300W and AFLW.} We perform the best on LR faces (bottom). Even with the proposed processing LR, while all others process  HR, it still is best (top).}
    {\footnotesize
    \begin{adjustbox}{max width=\linewidth}
        \begin{tabular}{m{13em}cccc}
            \toprule
            & \multicolumn{3}{c}{300W} & \multirow{2}{*}{AFLW} \tabularnewline 
            \cline{2-4} 
            & Common & Challenge & Full & \multirow{2}{*}{} \tabularnewline
            \toprule
            \vspace{0.2cm}
            SDM~\cite{xiong2013supervised} & 5.57 & 15.40 & 7.52 & 5.43 \tabularnewline
            LBF~\cite{ren2014face} & 4.95 & 11.98 & 6.32 & 4.25 \tabularnewline
            CFSS~\cite{zhu2015face} & 4.73 & 9.98 & 5.76 & 3.92 \tabularnewline
            MDM~\cite{trigeorgis2016mnemonic} & 4.83 & 10.14 & 5.88 & - \tabularnewline
            Two-stage~\cite{lv2017deep} & 4.36 & 7.56 & 4.99 & 2.17 \tabularnewline
            RCSR~\cite{wang2018recurrentpami} & 4.01 & 8.58 & 4.90 & - \tabularnewline
            CPM+SBR~\cite{dong2018supervision} & 3.28 & 7.58 & 4.10 & 2.14 \tabularnewline
            JASRNet (Ours) & \textbf{3.20} & \textbf{7.44} & \textbf{4.03} & \textbf{2.03} \tabularnewline 
            \midrule
            \midrule
            \vspace{0.20cm}
            SuperFAN~\cite{bulat2018superfan} & 5.60 & 10.47 & 6.55 & 3.774 \tabularnewline
            FSRNet~\cite{chen2018fsrnet} & 5.42& 10.76 & 6.46 & - - \tabularnewline
            CPM+SBR~\cite{dong2018supervision} & 5.42 & 10.65 & 6.45 & 3.87\tabularnewline
            JASRNet (Ours) & \textbf{4.60} & \textbf{8.10} & \textbf{5.29} & \textbf{3.35} \tabularnewline 
            \bottomrule
        \end{tabular}
    \label{tab:face-comparisons}
    \end{adjustbox}
    }
\end{table}

\begin{table}
    \centering
    \caption{\textbf{Quantitative comparisons on LFW.} Performance was measured using verification accuracy (ACC), PSNR, and SSIM. The number of parameters is also listed here.}
    {\footnotesize
    \begin{adjustbox}{max width=\linewidth}
        \begin{tabular}{ccccc}
            \toprule
             & ACC($\%$) & PSNR & SSIM & Param.\\\midrule 
            HR  & 99.33 & -- & -- & -- \\
            Bicubic &  79.50 & 25.28 & 0.736 & -- \\ 
            FSRNet~\cite{chen2018fsrnet}  & 83.75 & 26.63 & 0.800  & 27.14M\\
            SuperFAN (Bulat et al. 2018) & 84.08 & 26.83 & 0.808  & 26.41M\\
            Ours & \textbf{86.86} & \textbf{27.30} & \textbf{0.818} & 18.96M\\
            \bottomrule
        \end{tabular}
    \label{tab:compLFW}
    \end{adjustbox}
    }
\end{table}

\noindent\textbf{Face super-resolution results}.
We compared with methods used for SISR (\ie VDSR~\cite{kim2016accurate}, SRRes~\cite{ledig2017photo}, and EDSR~\cite{lim2017enhanced}), as well as  methods for face SR (\ie URDGN~\cite{yu2016ultra}, TDAE~\cite{yu2017hallucinating}, SuperFAN~\cite{bulat2018superfan} and FSRNet~\cite{chen2018fsrnet}). For a fair comparison, we retrained aforementioned models with the same training and testing data used in the respective experiment. Qualitative comparisons clearly show that the proposed JASRNet recovers HR images with relatively more details (\ie sharper edges, more accurate facial component shapes and textures), while other methods tend to produce face images with more blur and inaccuracies (see Figure \ref{fig:SRsmpImg}). 
Quantitative results for face SR are shown in Table~\ref{tab:resultsSR}. 
The proposed model achieved the highest PSNR and SSIM on 300W and HELEN dataset. Since some methods only support an upscaling factor of $\times$4, we added an additional upscaling module ($\times2$) to get the equivalent factor of $\times$8. For this, we incorporated the commonly used pixel shuffle followed by a convolutional layer~\cite{shi2016real}.

\noindent\textbf{Face alignment results}.
We present face alignment results for 300W and AFLW dataset with LR image size of $16\times16$ and $64\times64$ separately. The results are summarized in Table \ref{tab:face-comparisons}. First, we compare the results of $16\times16$ LR images (see bottom part of Table \ref{tab:face-comparisons}). Since only a few works address the tiny face (\ie $16\times16$) alignment problem, we only compare the performance of proposed models with SuperFAN, FSRNet, and another state-of-the-art method CPM+SBR~\cite{dong2018supervision}. Noticed that CPM+SBR is applied on super-resolved images using bicubic interpolation. Compared with other state-of-art methods, we show a large improvement for landmark localization on tiny faces.

Furthermore, we present results of JASRNet on faces with a resolution of $64\times64$ (see Table \ref{tab:face-comparisons} (top)). Note that existing methods detect landmarks on HR (\ie $256\times256$) images. Still, the proposed framework is comparable for landmark localization on LR images with the others on HR.

\begin{table*}[t]
\centering
\caption{\textbf{Ablation study.} To highlight the effectiveness of feature fusion and joint training.}
\label{tab:ablationStudy}
{\footnotesize
   \begin{adjustbox}{max width=\linewidth}
    \begin{tabular}{l|cc|cc|c}\toprule
    & \rotatebox{0}{baseline} & \rotatebox{0}{+feature fusion} & \rotatebox{0}{joint training}& \rotatebox{0}{+feature fusion} & \rotatebox{0}{JASRNet}\tabularnewline
    & (BL) & (BL\_F) & (JT) & (JT\_F) &  (ours)  
    \tabularnewline\midrule
Super Resolution       &   23.41    &   23.50    &     23.55     &   23.58   &   23.69        \tabularnewline
Face Alignment       &  5.71     &    5.70   &     5.34    &     5.34  &     5.26      \tabularnewline
\bottomrule
\end{tabular}
 \end{adjustbox}
}
\end{table*}


\begin{table}
    \centering
    \caption{\textbf{Baseline variations of the proposed JASRNet.} Trained and tested on 300W.}
    {\footnotesize
    \begin{adjustbox}{max width=\linewidth}
        \begin{tabular}{cccc}
            \toprule
             & \thead{Super Resolution \\ (PSNR)} & \thead{Face Alignment \\ (NMSE)} & \# of Param. \\\midrule
            Concat   & 23.57 & 5.42 & --\\
            Adding & 23.69 & 5.29 & --\\
            \midrule
            One\_stages  & 23.61 &  5.44 & 16.69M\\
            two\_stages & 23.62 &  5.36 & 17.83M\\
            Res\_16 & 23.62  & 5.41  & 14.46M\\
            Res\_32 & 23.69  & 5.29 & 18.96M\\
            \bottomrule
        \end{tabular}
    \label{tab:ablationStudyBaseline}
    \end{adjustbox}
    }
\end{table}

\noindent\textbf{Comparison on both tasks}.
To the best of our knowledge, FSRNet~\cite{chen2018fsrnet} and SuperFAN~\cite{bulat2018superfan} were the only attempts that reported results on both tasks (\ie SR and face alignment).
Thus, we compared results of both tasks with these two methods. Since one of the primary tasks for ``enhancing'' faces is to improve facial recognition capabilities, we also measured face verification performance on the super-resolved images. Additionally, the number of parameters used in each model is listed in Table~\ref{tab:compLFW}.
In this section, models were trained on the 300W training set, and tested on the 300W test set and the entire LFW dataset. The SR and alignment results for 300W test set are shown in Table~\ref{tab:resultsSR} and \ref{tab:face-comparisons}, respectively. As for LFW dataset, the results for SR and facial recognition are listed in Table \ref{tab:compLFW}. Performance was measured using verification accuracy (ACC), PSNR, and SSIM. We did not include LFW in the test for landmark localization since it does not support the 68 landmarks used as prior knowledge in all three methods. 
Thus, we show that our JASRNet significantly outperforms SuperFAN and FSRNet in face SR and landmark localization (see Table \ref{tab:resultsSR}, \ref{tab:face-comparisons}, and \ref{tab:compLFW}). Qualitatively, the proposed method also produces more accurate landmark estimations for alignment task and much more detailed appearances and texture for SR task than the other two methods (see Figure \ref{fig:resultsOnBothTasks}, \ref{fig:SRsmpImg}). Note that our model also have less parameters than SuperFAN and FSRNet  (see Table \ref{tab:compLFW}).

\subsection{Ablation study}\label{sec:ablation}
We next measured the contributions of feature fusion, joint training, and the long skip connection. Table~\ref{tab:ablationStudy} lists the four additional variants used. Baseline (BL) only consisted of an encoder, a feature extraction module, and either a SR or alignment module. In other words, the BL omitted the feature fusion at the intermediate layers, removed the long skip connection, and was only able to handle a single task per pass (\ie either SR or face alignment, but not both). BL\_F is BL with feature fusion. Joint training (JT) net was conducted by aggregating both task-specific modules to the baseline, and JT with feature fusion is JT\_F. Finally, JT\_F with long skip connection forms the proposed JASRNet. The training set used in this section is 300W. Note that our baseline model has even better performance while less parameters than SuperFAN~\cite{bulat2018superfan}.
Reasons are three-fold: 1) batch normalization omitted in layers of residual blocks to speed up training and boost performance; 2) Pixel shuffle layers~\cite{shi2016real} used in reconstruction module instead of deconvolutional, which is used in SuperFAN; 3) Two independent modules are used in SuperFAN, i.e., SR and face alignment are handled separately. This yields redundant feature maps and, hence, degrades performance. 

\noindent\textbf{Effects of the feature fusion}.
Fusing the features at the intermediate layers yields richer, and more efficient feature representations for SR, with BL\_F and JT\_F outperforming BL and JT, respectively, in SR (see Table~\ref{tab:ablationStudy}). 
However, feature fusion has less impact on face alignment. This is because SR uses both low and high frequency information to recover HR from LR images, while landmark localization is mostly dependent on the high frequency content.


\noindent\textbf{Effects of joint-task mechanism}.
To highlight the importance of training the two tasks jointly, we compared JT to BL and JT\_F to BL\_F (see Table~\ref{tab:ablationStudy}). Results for both tasks (\ie SR and face alignment) show that joint-task variants (\ie JT and JT\_F) significantly outperform BL and BL\_F, respectfully. This validates that the joint training, in itself, contributes to the state-of-the-art performance of JASRNet.

\noindent\textbf{Effects of long skip connection}.
The impact of the long skip connection is evident by the results: JASRNet, which is JT\_F with the added skip connection, outperforms all others in both SR and landmark localization. The impact for SR stems from the skip connection forcing the network to encode sharper and more precise edges in the feature representation, as expected. However, the boosted accuracy for face alignment was less expected, yet supporting of the narrative: we believe the shared features for SR and face alignment yield additional information that complements both tasks.

\noindent\textbf{Baseline variations}.
We also show the variations of the vanilla baseline for insights on the effects of different fusion methods (\ie concatenation vs element-wise addition), the number of residual blocks in the feature extraction module (\ie 16 vs 32), and the number of stages in the face alignment module (\ie 1 vs 2). Table \ref{tab:ablationStudyBaseline} lists results for different settings. Clearly, element-wise addition is better for the feature fusion module in our model. Also, more residual blocks and stages improves the performance. Thus, the deeper structure and, thus, the higher capacity captures more information for the SR and face alignment tasks: as the network grows so does its potential to learn.

\section{Conclusion}
We proposed a JASRNet to exploit the maximum amount of information from tiny face images when simultaneously addressing alignment and super-resolution tasks.
Extensive experiments demonstrated the proposed significantly outperforms previous state-of-the-art in SR by recovering sharper edges (\ie finer details) from HR faces.
We also show large improvements for landmark localization of tiny faces (\ie $16\times16$). Furthermore, the proposed framework yields comparable results for landmark localization on faces of lower-resolution (\ie $64\times64$) to existing methods on higher-resolution (\ie $256\times256$).

\bibliographystyle{aaai}
\bibliography{aaai}

\end{document}